\documentclass[a4paper,english]{article}
\usepackage[latin9]{inputenc}
\usepackage{color}
\usepackage{babel}
\usepackage{array}
\usepackage{multirow}
\usepackage{amstext}
\usepackage{graphicx}

\makeatletter

\pdfpageheight\paperheight
\pdfpagewidth\paperwidth

\providecommand{\tabularnewline}{\\}

\usepackage{a4wide}

\usepackage{xspace}
\DeclareRobustCommand\onedot{\futurelet\@let@token\@onedot}
\def\@onedot{\ifx\@let@token.\else.\null\fi\xspace}

\def\eg{\emph{e.g}\onedot} 
\def\ie{\emph{i.e}\onedot}

\def\etal{\emph{et al}\onedot}

\@ifundefined{showcaptionsetup}{}{%
 \PassOptionsToPackage{caption=false}{subfig}}
\usepackage{subfig}
\makeatother

\begin{document}

\global\long\def\product{\cdot}


\global\long\def\centerTarget{C}

\global\long\def\ourModel{\text{HitNet}}


\global\long\def\cardinalityOfClasses{K}

\global\long\def\classIndex{k}


\global\long\def\loss{L}

\global\long\def\output{y}

\global\long\def\true#1{#1_{\text{true}}}

\global\long\def\indexedTrue#1#2{#1_{\text{true},\,#2}}

\global\long\def\predicted#1{#1_{\text{pred}}}

\global\long\def\indexedPredicted#1#2{#1_{\text{pred},\,#2}}


\global\long\def\matrixOfHinton{M}

\global\long\def\mHinton{m}

\global\long\def\row{x}

\global\long\def\dimension{n}


\global\long\def\trainingImage{X}

\global\long\def\reconstructed#1{#1_{\text{rec}}}

\global\long\def\modified#1{#1_{\text{mod}}}

\global\long\def\LTwo{L^{2}}

\global\long\def\length#1{\left\Vert #1\right\Vert }


\global\long\def\FreeShot{\text{F}}

\global\long\def\numberOfFSClasses{N_{c}}

\global\long\def\Heaviside#1{H(#1)}

\title{HitNet: a neural network with capsules embedded in a Hit-or-Miss
layer, extended with hybrid data augmentation and ghost capsules}

\author{Adrien Deliège, Anthony Cioppa, Marc Van Droogenbroeck}
\maketitle
\begin{abstract}
Neural networks designed for the task of classification have become
a commodity in recent years. Many works target the development of
better networks, which results in a complexification of their architectures
with more layers, multiple sub-networks, or even the combination of
multiple classifiers. In this paper, we show how to redesign a simple
network to reach excellent performances, which are better than the
results reproduced with CapsNet on several datasets, by replacing
a layer with a Hit-or-Miss layer. This layer contains activated vectors,
called capsules, that we train to hit or miss a central capsule by
tailoring a specific centripetal loss function. We also show how our
network, named $\ourModel$, is capable of synthesizing a representative
sample of the images of a given class by including a reconstruction
network. This possibility allows to develop a data augmentation step
combining information from the data space and the feature space, resulting
in a hybrid data augmentation process. In addition, we introduce the
possibility for $\ourModel$, to adopt an alternative to the true
target when needed by using the new concept of ghost capsules, which
is used here to detect potentially mislabeled images in the training
data.
\end{abstract}

\section{Introduction}

\begin{figure}[b]
\begin{centering}
\includegraphics[width=0.99\columnwidth]{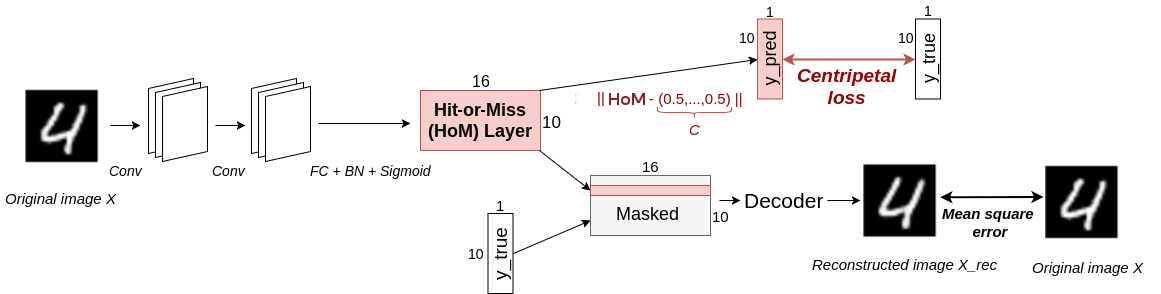}
\par\end{centering}
\caption{Graphical representation of the structure of our new network, named
$\protect\ourModel$. Our contributions are highlighted in red, and
comprise a new Hit-or-Miss layer, a centripetal loss, prototypes that
can be built with the decoder, and ghost capsules that can be embedded
in the HoM layer. Source code is available at \protect\href{http://www.telecom.ulg.ac.be/hitnet}{http://www.telecom.ulg.ac.be/hitnet}.\label{fig:HitNet-structure}}
\end{figure}

Convolutional neural networks (CNNs) have become an omnipresent tool
for image classification and have been revolutionizing the field of
computer vision for the last few years. With the emergence of complex
tasks such as ImageNet classification~\cite{Deng2009ImageNet}, the
networks have grown bigger and deeper while regularly featuring new
layers and other extensions. However, CNNs are not intrinsically viewpoint-invariant,
meaning that the spatial relations between different features are
generally not preserved when using CNNs. Therefore, some models were
designed in the spirit of increasing their representational power
by encapsulating information in activated vectors called capsules,
a notion introduced by Hinton in~\cite{Hinton2011Transforming}.

Recent advances on capsules are presented in~\cite{Sabour2017Dynamic},
in which Sabour~\etal mainly focus on MNIST digits classification~\cite{LeCun2001Gradient}.
For that purpose, they develop CapsNet, a CNN that shows major changes
compared to conventional CNNs. As described in~\cite{Sabour2017Dynamic},
``a capsule is a group of neurons whose activity vector represents
the instantiation parameters of a specific type of entity such as
an object or an object part.'' Hence, the concept of capsule somehow
adds a (geometrical) dimension to the ``capsuled'' layers, which
is meant to contain richer information about the features captured
by the network than in conventional feature maps. The transfer of
information from the capsules of a layer to the capsules of the next
layer is learned through a dynamic routing mechanism~\cite{Hinton2018Matrix,Sabour2017Dynamic}.
The length of the capsules of the last layer, called DigitCaps, is
used to produce a prediction vector whose entries are in the $[0,1]$
range thanks to an orientation-preserving squashing activation applied
beforehand to each capsule, and which encodes the likelihood of the
existence of each digit on the input image. This prediction vector
is evaluated through a ``margin loss'' that displays similarities
with the squared Hinge loss. In an encoder-decoder spirit, the capsules
of DigitCaps can be fed to a decoder sub-network that aims at reconstructing
the initial image, which confers the capsules a natural interpretation
of the features that they encoded.State-of-the-art results are reported
by Sabour \etal in~\cite{Sabour2017Dynamic} on MNIST dataset. Other
experiments carried out on affNIST~\cite{Tieleman2013AffNIST}, multiMNIST~\cite{Sabour2017Dynamic},
SVHN~\cite{Netzer2011Reading}, smallNORB~\cite{LeCun2004Learning}
and CIFAR10~\cite{Krizhevsky2009Learning} (with an ensemble of 7
networks) show promising results as well. Unfortunately, current implementations
of CapsNet with dynamic routing are considerably slower than conventionnal
CNNs, which is a major drawback of this process. 

Since the publication of~\cite{Sabour2017Dynamic}, several works
have been conducted to improve CapsNet's speed and structure (\cite{Bahadori2018Spectral,Hinton2018Matrix,Rawlinson2018Sparse,Wang2018AnOptimization})
and to apply it to more complex data (\cite{Afshar2018Brain,Li2018TheRecognition,ONeill2018Siamese})
and various tasks (\cite{Liu2018Object} for localization, \cite{Lalonde2018Capsules}
for segmentation, \cite{Wang2018AnAttentionBased} for hypernymy detection,
\cite{Andersen2018DeepReinforcement} for reinforcement learning).
However, it appears that all the attempts (\eg \cite{Guo2017CapsNetKeras,Liao2018CapsNet,Nair2018Pushing,Shin2018CapsNetTensorFlow})
to reproduce the results provided in~\cite{Sabour2017Dynamic} failed
to reach the performances reported by Sabour \etal. 

The first part of this work is devoted to the construction of a neural
network, named \emph{HitNet}, that uses the capsule approach only
in one layer, called \emph{Hit-or-Miss layer} (HoM, the counterpart
of DigitCaps) and that provides fast and repeatedly better performances
than those reported in \cite{Guo2017CapsNetKeras,Liao2018CapsNet,Nair2018Pushing,Shin2018CapsNetTensorFlow}
with CapsNet. We also provide its associated loss, that we call \emph{centripetal
loss }(counterpart of the margin loss).

The strong representational behavior expected from CapsNet allows
to perform the joint multi-task of classification and reconstruction.
It is thus possible to take advantage of capsules to capture the natural
variations of important class-specific features of the training data,
as illustrated in~\cite{Sabour2017Dynamic}. By browsing through
the space of features and using the decoder appropriately, it is thus
possible to perform data generation and data augmentation. Augmenting
the data is recognized as a powerful way to prevent overfitting and
increase the ability of the network to generalize to unseen data at
test time, which leads to better classification results~\cite{Perez2017TheEffectiveness}.
This process is often applied entirely either in the data space or
in the feature space~\cite{Wong2016Understanding}. As a second part,
we present a way of using the capsules of HoM to derive a hybrid data
augmentation algorithm that relies on both real data and synthetic
feature-based data by introducing the notion of \emph{prototype},
a class representative learned indirectly by the decoder.

Not only do we use capsules for data-driven applications such as data
generation and data augmentation, but we can also use them to define
new notions that serve novel purposes. The one that we want to highlight
as third part is the possibility to allow $\ourModel$ to adopt an
alternative choice to the true class when needed, through the notion
of \emph{ghost capsule }that we embed in HoM\emph{. }More specifically,
in this work, we show how to use ghost capsules to analyze the training
set and to detect potentially mislabeled training images, which is
often eluded in practice despite being of paramount importance.

In short, our contributions are threefold:
\begin{enumerate}
\item We develop $\ourModel$, a neural network that uses capsules in a
new way through a Hit-or-Miss layer and a centripetal loss, and we
demonstrate the superiority of $\ourModel$ over the results reproduced
by other authors with CapsNet.
\item We derive a hybrid data space and feature space data augmentation
process via the capsules of HoM and prototypes.
\item We provide a way for $\ourModel$ to identify another plausible class
for the training images if necessary with the new notion of ghost
capsules. We exemplify this notion by detecting potentially mislabeled
training images, or training images that may need two labels.
\end{enumerate}
These contributions are described in that order in Section~\ref{sec:HitNet-and-beyond},
then tested in Section~\ref{sec:Experiments-and-results}.

\paragraph*{\emph{}}

\section{$\protect\ourModel$: rethinking DigitCaps and beyond\label{sec:HitNet-and-beyond}}

$\ourModel$ essentially introduces a new layer, the Hit-or-Miss layer,
that is universal enough to be used in many different networks. $\ourModel$
as presented hereafter is thus an instance of a shallow network that
hosts this HoM layer and illustrates its potential.

\subsection{Introducing hits, misses, the centripetal approach, and the Hit-or-Miss
layer\label{subsec:Centripetal-loss-and-HitNet}}

 In the case of CapsNet, large activated values are expected from
the capsule of DigitCaps corresponding to the true class of a given
image, similarly to usual networks. From a geometrical perspective
in the feature space, this results in a capsule that can be seen as
a point that the network is trained to push far from the center of
the unit hypersphere, in which it ends up thanks to the squashing
activation function. We qualify such an approach as ``centrifugal''.
In that case, a first possible issue is that one has no control on
the part(s) of the sphere that will be targeted by CapsNet, and a
second one is that the capsules of two images of the same class might
be located far from each other (\cite{Shahroudnejad2018Improved,Zhang2018CapProNet}),
which are two debatable behaviors. 

To solve these issues, we hypothesize that all the images of a given
class share some class-specific features and that this assumption
should also manifest through their respective capsules. Hence, given
an input image, we impose that $\ourModel$ targets the center of
the feature space to which the capsule of the true class belongs,
so that it corresponds to what we call a \emph{hit}. The capsules
related to the other classes have thus to be sent far from the center
of their respective feature spaces, which corresponds to what we call
a\emph{ miss}. Our point of view is thus the opposite of Sabour \etal's;
instead, we have a\emph{ centripetal approach} with respect to the
true class.

The squashing activation function induces a dependency between the
features of a capsule of DigitCaps, in the sense that their values
are conditioned by the overall length of the capsule. If one feature
of a capsule has a large value, then the squashing prevents the other
features of that capsule to take large values as well; alternatively,
if the network wishes to activate many features in a capsule, then
none of them will be able to have a large value. None of these two
cases fit with the perspective of providing strong activations for
several representative features as desired in Sabour \etal. Besides,
the orientation of the capsules, preserved with the squashing activation,
is not used explicitly for the classification; preserving the orientation
might thus be a superfluous constraint.

Therefore, we replace this squashing activation by a BatchNormalization
(BN,~\cite{Ioffe2015BatchNormalization}) followed by a conventional
sigmoid activation function applied element-wise. We obtain a layer
composed of capsules as well that we call the\emph{ Hit-or-Miss} (HoM)
layer, which is $\ourModel$'s counterpart of DigitCaps. Consequently,
all the features obtained in HoM's capsules can span the $[0,1]$
range and they can reach any value in this interval independently
of the other features. The feature spaces in which the capsules of
HoM lie are thus unit hypercubes.

\subsubsection*{Defining the centripetal loss}

\begin{figure}
\subfloat[\label{fig:loss-hit}]{\includegraphics[width=0.47\textwidth]{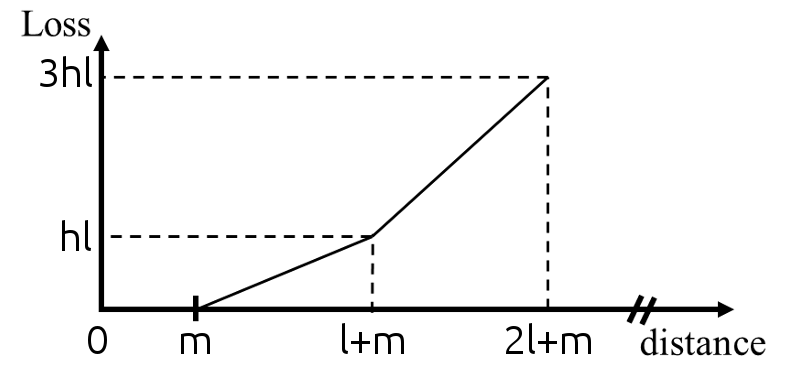}}\hfill{}\subfloat[\label{fig:loss-miss}]{\includegraphics[width=0.47\textwidth]{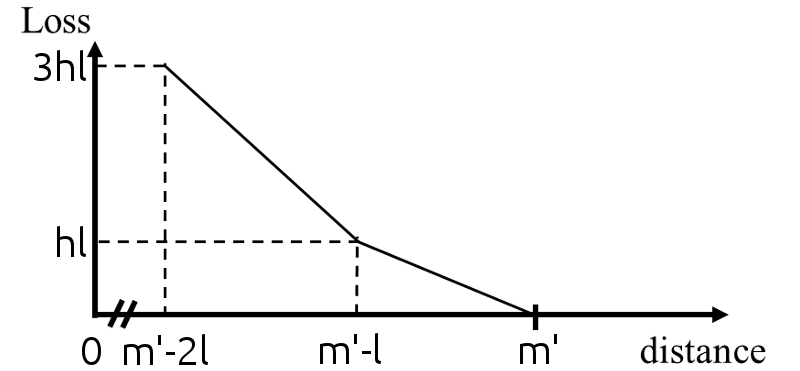}}

\subfloat[\label{fig:Target-to-hit}]{\includegraphics[width=0.47\textwidth]{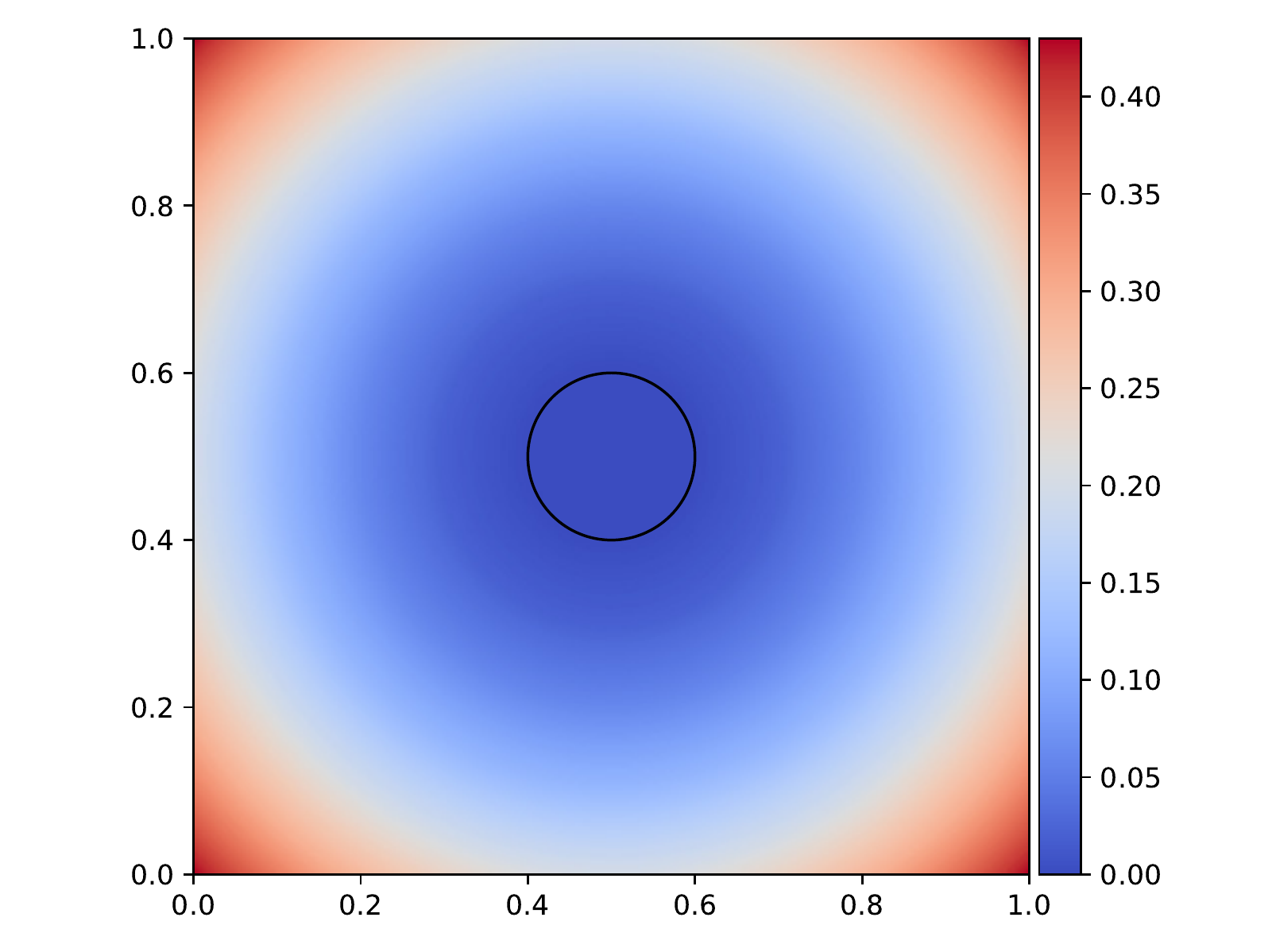}}\hfill{}\subfloat[\label{fig:Target-to-miss}]{\includegraphics[width=0.47\textwidth]{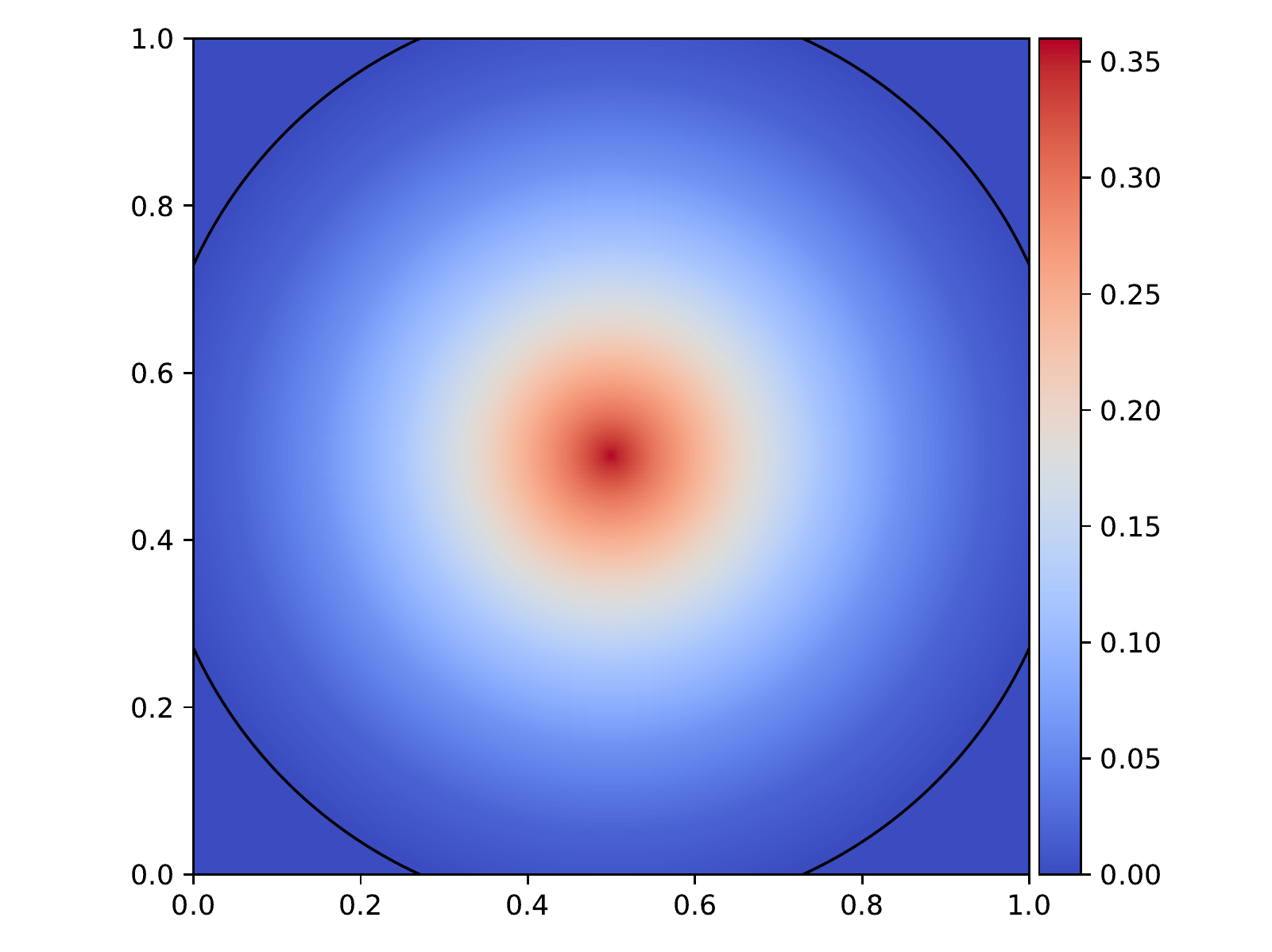}}\caption{Representation of the centripetal loss given in Equation~\ref{eq:steploss1}
as function of the distance of the capsule of the true class (a) and
of the other classes (b) from $\protect\centerTarget$. Then, visualization
of the centripetal loss in the 2-dimensional case ($\protect\dimension=2$).
The loss associated with the capsule of the true class is given by
plot (c). The loss-free hit zone is the area within the black circle,
with radius $m$. The loss generated by the other capsules is given
by plot (d). The loss-free miss zone is the area outside the black
circle, with radius $m'$. \label{fig:mylosses} }
\end{figure}

Given the use of the element-wise sigmoid activation, the centers
of the reshaped target spaces are, for each of them, the \emph{central
capsules} $\centerTarget:(0.5,\,\ldots,\,0.5)$. The $\classIndex$-th
component of the prediction vector $\predicted y$ of $\ourModel$,
denoted $\indexedPredicted{\output}{\classIndex}$, is given by the
Euclidean distance between the $\classIndex$-th capsule of HoM and
$\centerTarget$:
\begin{equation}
\indexedPredicted{\output}{\classIndex}=||\text{HoM}_{\classIndex}-\centerTarget||.\label{eq:ourpred}
\end{equation}
To give a tractable form to the notions of hits, misses, centripetal
approach described above and justify HoM's name, we design a custom
centripetal loss function with the following requirements:
\begin{enumerate}
\item The loss generated by each capsule of HoM has to be independent of
the other capsules. We thus get rid of any probabilistic notion during
the training.
\item The capsule of the true class does not generate any loss when belonging
to a close isotropic neighborhood of $\centerTarget$, which defines
the \emph{hit zone}. Outside that neighborhood, it generates a loss
increasing with its distance to $\centerTarget$. The capsules related
to the remaining classes generate a loss decreasing with their distance
to $\centerTarget$ inside a wide neighborhood of $\centerTarget$
and do not generate any loss outside that neighborhood, which is the
\emph{miss zone}. These loss-free zones are imposed to stop penalizing
capsules that are already sufficiently close (if associated with the
true class) or far (if associated with the other classes) from $\centerTarget$
in their respective feature space.
\item The gradient of the loss with respect to $\indexedPredicted{\output}k$
cannot go to zero when the corresponding capsule approaches the loss-free
zones defined in requirement 2. To guarantee this behavior, we impose
a constant gradient around these zones. This is imposed to help the
network make hits and misses.
\item For the sake of consistency with requirement 3, we impose piecewise
constant gradients with respect to $\indexedPredicted{\output}k$,
which thus defines natural bins around $\centerTarget$, as the rings
of archery targets, in which the gradient is constant.
\end{enumerate}
All these elements contribute to define a loss which is a piecewise
linear function of the predictions and which is \emph{centripetal
with respect to the capsule of the true class}. We thus call it our
\emph{centripetal loss}. Its derivative with respect to $\indexedPredicted{\output}k$
is a staircase-like function, which goes up when $k$ is the index
of the true class (see Figure~\ref{fig:loss-hit}) and goes down
otherwise (see Figure~\ref{fig:loss-miss}). A generic analytic
formula of a function of a variable $x$, whose derivative is an increasing
staircase-like function where the steps have length $l$ and height
$h$ and vanish on $[0,m]$ is mathematically given by:

\begin{equation}
L_{l,h,m}(x)=H\{x-m\}\,(f+1)\,h\,(x-m-0.5\,f\,l),\label{eq:steploss1}
\end{equation}

where $H\{.\}$ denotes the Heaviside step function and $f=\left\lfloor \frac{x-m}{l}\right\rfloor $
($\left\lfloor .\right\rfloor $ is the floor function). Hence the
loss generated by the capsule of the true class is given by $L_{l,h,m}(\indexedPredicted{\output}k)$,
where $k$ is the index of the true class. The loss generated by the
capsules of the other classes can be directly obtained from Equation~\ref{eq:steploss1}
as $L_{l',h',\sqrt{n}/2-m'}(\sqrt{n}/2-\indexedPredicted{\output}{k'})$
(for any index $k'$ of the other classes) if the steps have length
$l'$, height $h'$, vanish after $m'$ and if the capsules have $n$
components. The use of $\sqrt{n}/2$ originates from the fact that
the maximal distance between a capsule of HoM and $\centerTarget$
is given by $\sqrt{n}/2$ and thus the entries of $\predicted{\output}$
will always be in the interval $[0,\sqrt{n}/2]$. Consequently, the
centripetal loss of a given training image is given by

\begin{equation}
\loss=\sum_{\classIndex=1}^{\cardinalityOfClasses}\indexedTrue{\output}{\classIndex}\,L_{l,h,m}(\indexedPredicted{\output}{\classIndex})+\lambda(1-\indexedTrue{\output}{\classIndex})\,L_{l',h',\sqrt{n}/2-m'}(\sqrt{n}/2-\indexedPredicted{\output}{\classIndex})\label{eq:myloss-1}
\end{equation}
where $\cardinalityOfClasses$ is the number of classes, $\indexedTrue yk$
denotes the $\classIndex$-th component of the vector $\true y$,
and $\lambda$ is a down-weighting factor set as $0.5$ as in~\cite{Sabour2017Dynamic}.
The loss associated with the capsule of the true class and the loss
associated with the other capsules are represented in Figure~\ref{fig:mylosses}
in the case where $\dimension=2$.

\subsubsection*{Architecture of $\protect\ourModel$}

Basically, $\ourModel$ incorporates a HoM layer built upon feature
maps and used in pair with the centripetal loss. In our experiments,
we have adopted a shallow structure to obtain these feature maps to
highlight the benefits of the HoM layer. $\ourModel$'s complete architecture
is displayed in Figure~\ref{fig:HitNet-structure}. It is composed
of the following elements:
\begin{enumerate}
\item Two $9\times9$ (with strides (1,1) then (2,2)) convolutional layers
with 256 channels and ReLU activations, to obtain feature maps.
\item A fully connected layer to a $\cardinalityOfClasses\times\dimension$
matrix, followed by a BN and an element-wise sigmoid activation, which
produces HoM composed of $\cardinalityOfClasses$ capsules of size
$\dimension$.
\item The Euclidean distance with the central capsule $\centerTarget:(0.5,\,\ldots,\,0.5)$
is computed for each capsule of HoM, which gives the prediction vector
of the model $\predicted{\output}$.
\item All the capsules of HoM are masked (set to $0$) except the one related
to the true class (to the predicted class at test time), then they
are concatenated and sent to a decoder, which produces an output image
$\reconstructed{\trainingImage}$, that aims at reconstructing the
initial image. The decoder consists in two fully connected layers
of size $512$ and $1024$ with ReLU activations, and one fully connected
layer to a matrix with the same dimensions as the input image, with
a sigmoid activation (this is the same decoder as in~\cite{Sabour2017Dynamic}).
\end{enumerate}
If $\trainingImage$ is the initial image and $\true{\output}$ its
one-hot encoded label, then $\true{\output}$ and $\predicted{\output}$
produce a loss $L_{1}$ through the centripetal loss given by Equation~\ref{eq:myloss-1}
while $\trainingImage$ and $\reconstructed{\trainingImage}$ generate
a loss $L_{2}$ through the mean squared error. The final composite
loss associated with $X$ is given by $L=L_{1}+\alpha\,L_{2}$, where
$\alpha$ is set to $0.392$ (\cite{Guo2017CapsNetKeras,Sabour2017Dynamic}).
For the classification task, the label predicted by $\ourModel$ is
the index of the lowest entry of $\predicted y$. The hyperparameters
involved in $L_{1}$ are chosen as $l=l'=0.1$, $h=h'=0.2$, $m=0.1$,
$m'=0.9$, $n=16$ and $\lambda=0.5$ as in~\cite{Sabour2017Dynamic}.

\subsection{Beyond $\protect\ourModel$: Prototypes, data generation and hybrid
data augmentation\label{subsec:Prototypes,-data-generation-1}}

\subsubsection*{Prototypes}

The simultaneous use of HoM and a decoder offers new possibilities
in terms of image processing. It is essential to underline that in
our centripetal approach, we ensure that all the images of a given
class will have all the components of their capsule of that class
close to $0.5$. In other words, we regroup these capsules in a convex
space around $\centerTarget$. This central capsule $\centerTarget$
stands for a fixed point of reference, hence different from a centroid,
from which we measure the distance of the capsules of HoM; from the
network's point of view, $\centerTarget$ stands for a capsule of
reference from which we measure deformations.  In consequence, we
can use $\centerTarget$ instead of the capsule of a class of HoM,
zero out the other capsules and feed the result in the decoder: the
reconstructed image will correspond to the image that the network
considers as a canonical image of reference for that class, which
we call its \emph{prototype.}

\subsubsection*{Data generation}

After constructing the prototypes, we can slightly deform them to
induce variations in the reconstruction without being dependent on
any training image, just by feeding the decoder with a zeroed out
HoM plus one capsule in a neighborhood of $\centerTarget$. This allows
to identify what the features of HoM represent. For the same purpose,
Sabour \etal need to rely on a training image because the centrifugal
approach does not directly allows to build prototypes. In our case,
it is even possible to compute an approximate range in which the components
can be tweaked. If a sufficient amount of training data is available,
we can expect the individual features of the capsules of the true
classes to be approximately Gaussian distributed with mean $0.5$
and standard deviation $m/\sqrt{n}$\footnote{Comes from Equation~\ref{eq:ourpred}, with the hypothesis that all
the values of such a capsule differ from $0.5$ from roughly the same
amount.}, thus the $[0.5-2m/\sqrt{n},0.5+2m/\sqrt{n}]$ interval should provide
a satisfying overview of the physical interpretation embodied in a
given feature of HoM. The approximate knowledge of the distributions
also enables us to perform data generation, by sampling for instance
a Gaussian vector of size $n$, with mean $0.5$ and standard deviation
$m/\sqrt{n}$ , inserting it as a capsule in HoM, zeroing out the
other capsules and feeding the result in the decoder.

\subsubsection*{Hybrid data augmentation}

The capsules of HoM only capture the important features that allow
the network to identify the class of the images and to perform an
approximate reconstruction via the decoder. This implies that the
images produced by the decoder are not detailed enough to look realistic.
The details are lost in the process; generating them back is hard.
It is easier to use already existing details, \ie those of images
of the training set. We can thus set up a hybrid feature-based and
data-based data augmentation process:
\begin{itemize}
\item Take a training image $\trainingImage$ and feed it to a trained $\ourModel$
network.
\item Extract its HoM and modify the capsule corresponding to the class
of $\trainingImage$.
\item Reconstruct the image obtained from the initial capsule, $\reconstructed{\trainingImage}$,
and from the modified one, $\modified{\trainingImage}$.
\item The details of $\trainingImage$ are contained in $\trainingImage-\reconstructed{\trainingImage}$.
Thus the new (detailed) image is $\modified{\trainingImage}+\trainingImage-\reconstructed{\trainingImage}$.
Clip the values to ensure that the resulting image has values in the
appropriate range (\eg~{[}0,1{]}).
\end{itemize}

\subsection{Beyond $\protect\ourModel$: Ghost capsules\label{subsec:Ghost-capsules}}

One of the main assets of $\ourModel$ is the flexibility in the use
of the HoM layer. It can be easily exploited to perform different
tasks other than classification. In order to show an additional possibility
of $\ourModel$, we develop the notion of \emph{ghost} \emph{capsules,}
that can be integrated within the network. The use of ghost capsules
allows the network, for each training sample, to zero-out the loss
associated with a capsule related to a class that the network considers
as a reasonable prediction and that is different from the true class.
Several situations can benefit from this process. For example, it
can be used to assess the quality of the training set through the
detection of potentially mislabeled images in that set. Detecting
them is an important aspect since mislabeled training data pollute
the whole training by forcing the network to learn unnatural features,
or even mistakes; this constrains the network to memorize outliers
in dedicated neurons.

\subsubsection*{Definition of ``ghost capsule''}

In order to give the network the capacity to allow an alternative
to the labels provided, we introduce the notion of ``ghost capsule'',
which is associated with every image of the training set. The key
idea is the following: during the training phase, for each image,
instead of forcing the network to produce one capsule that makes a
hit and $\cardinalityOfClasses-1$ capsules that make misses, we demand
one hit for the capsule of the true class and $\cardinalityOfClasses-2$
misses for as many capsules of the other classes; the remaining capsule
is the so-called \emph{``ghost capsule''}, denoted by GC hereafter.
By ``capsule of the true class'' of an image, we mean the capsule
associated with the class corresponding to the label provided by the
annotator with the image. The GC is the capsule of the HoM layer which
is the closest to $\centerTarget$ among the $\cardinalityOfClasses-1$
capsules not corresponding to the true class. The loss associated
with the GC is zeroed out, hence the GC is not forced to make a hit
nor a miss, and it is not involved in the update of the weights from
one batch to the next; it is essentially invisible to the loss in
the back-propagation, hence its name.

From an implementation point of view, training $\ourModel$ with a
GC per image is similar to training it without GC; only the centripetal
loss needs to be adjusted. Given a training image, its one-hot encoded
label $\true{\output}$ and the output vector of the network $\predicted{\output}$,
the centripetal loss initially defined by Equation~\ref{eq:myloss-1}
formally becomes

\begin{equation}
\loss_{\text{ghost}}=\sum_{\classIndex=1}^{\cardinalityOfClasses}\indexedTrue{\output}{\classIndex}\,L_{l,h,m}(\indexedPredicted{\output}{\classIndex})+\lambda(1-{\color{red}\indexedTrue{\tilde{\output}}{\classIndex}})\,L_{l',h',\sqrt{n}/2-m'}(\sqrt{n}/2-\indexedPredicted{\output}{\classIndex}),\label{eq:lossghost}
\end{equation}
where ${\color{red}\indexedTrue{\tilde{\output}}{\classIndex}}=1$
if $\classIndex$ is the true class index or the GC class index, and
${\color{red}\indexedTrue{\tilde{\output}}{\classIndex}}=0$ otherwise.

Two important characteristics are thus associated with a GC: its class,
which is always one of the $\cardinalityOfClasses-1$ classes not
corresponding to the true class of the sample, and its distance with
$\centerTarget$. The GC class of an image is obtained in a deterministic
way, it is not a Dropout (\cite{Srivastava2014Dropout}) nor a learned
Dropout variant (as \eg \cite{Lee2017DropMax}). Besides, this class
is likely to change from one epoch to the next in the early stages
of the training, until the network decides what choice is the best.
Ideally, a GC will make a hit when its class is a plausible alternative
to the true class or if the image deserves an extra label, and will
make a miss otherwise. The evolution of a GC during the training is
dictated by the overall evolution of the network. 

Subsequently, in the situations described above, at the end of the
training, mislabeled or confusing images should actually display two
hits: one hit for the capsule corresponding to the true class since
the network was forced to make that hit, and one hit for the capsule
corresponding to the other plausible class since the network was not
told to push this capsule towards the miss zone and since the image
displays the features needed to identify this alternate class. Looking
at the images with a GC in the hit zone at the end of the training
allows to detect the images for which the network suspects an error
in the label, or which images possibly deserve two labels.

\section{Experiments and results\label{sec:Experiments-and-results}}

In this section, we present some experiments and the results obtained
with $\ourModel$. The structure of the section mirrors that of Section~\ref{sec:HitNet-and-beyond},
\ie it is divided in three parts, which directly correspond to the
three parts of Section~\ref{sec:HitNet-and-beyond}: (1) general
performances of $\ourModel$, (2) use of the decoder, and (3) use
of ghost capsules.

\subsection{Classification results of $\protect\ourModel$}

Hereafter, we report the results obtained with $\ourModel$. First,
we compare the performances of $\ourModel$ on MNIST classification
task with baseline models to show that $\ourModel$ produces repeatedly
close to state-of-the-art performances. Then, we compare the performances
of $\ourModel$ with reproduced experiments with CapsNet on several
datasets.

\subsubsection*{Description of the networks used for comparison}

We compare the performances of $\ourModel$ to three other networks
for the MNIST digits classification task. For the sake of a fair comparison,
a structure similar to $\ourModel$ is used as much as possible for
these networks. First, they are made of two $9\times9$ convolutional
layers with 256 channels (with strides (1,1) then (2,2)) and ReLU
activations as for $\ourModel$. Then, the first network is N1:
\begin{itemize}
\item N1 (baseline model, conventional CNN) has a fully connected layer
to a vector of dimension $10$, then BN and Softmax activation, and
is evaluated with the usual categorical cross-entropy loss. No decoder
is used.
\end{itemize}
The two other networks, noted N2 and N2b, have a fully connected layer
to a $10\times16$ matrix, followed by a BN layer as N1 and $\ourModel$,
then
\begin{itemize}
\item N2 (CapsNet-like model) has a squashing activation. The Euclidean
distance with $O:(0,\,\ldots,\,0)$ is computed for each capsule,
which gives the output vector of the model $\predicted{\output}$.
The margin loss (centrifugal) of~\cite{Sabour2017Dynamic} is used;
\item N2b has a sigmoid activation. The Euclidean distance with $\centerTarget:(0.5,\,\ldots,\,0.5)$
is computed for each capsule, which gives the output vector of the
model $\predicted{\output}$. The margin loss (centrifugal) of~\cite{Sabour2017Dynamic}
is used. 
\end{itemize}
Let us recall that, compared to N2 and N2b, $\ourModel$ has a sigmoid
activation, which produces the capsules of HoM. The $\LTwo$ distance
with $\centerTarget:(0.5,\,\ldots,\,0.5)$ is computed for each capsule,
which gives the output vector of the model $\predicted{\output}$,
and the centripetal loss given by Equation~\ref{eq:myloss-1} is
used. Network N2b is tested to show the benefits of the centripetal
approach of $\ourModel$ over the centrifugal one, regardless of the
squashing or sigmoid activations. Also, during the training phase,
the decoder used in $\ourModel$ is also used with N2 and N2b. 

\subsubsection*{Classification results on MNIST}

Each network is trained $20$ times during $250$ epochs with the
Adam optimizer with a constant learning rate of $0.001$, with batches
of $128$ images. The images of a batch are randomly shifted of up
to $2$ pixels in each direction (left, right, top, bottom) with zero
padding as in~\cite{Sabour2017Dynamic}. The metrics presented here
for a given network are the average metrics over the $20$ runs of
that network calculated on the MNIST test set. The learning rate
is kept constant to remove its possible influence on the results,
which may differ from one network structure to another. Besides, this
lead us to evaluate the ``natural'' convergence of the networks
since the convergence is not forced by a decreasing learning rate
mechanism. To our knowledge, this practice is not common but should
be used to properly analyze the natural convergence of a network.

\begin{figure}
\centering{}\includegraphics[width=0.8\textwidth]{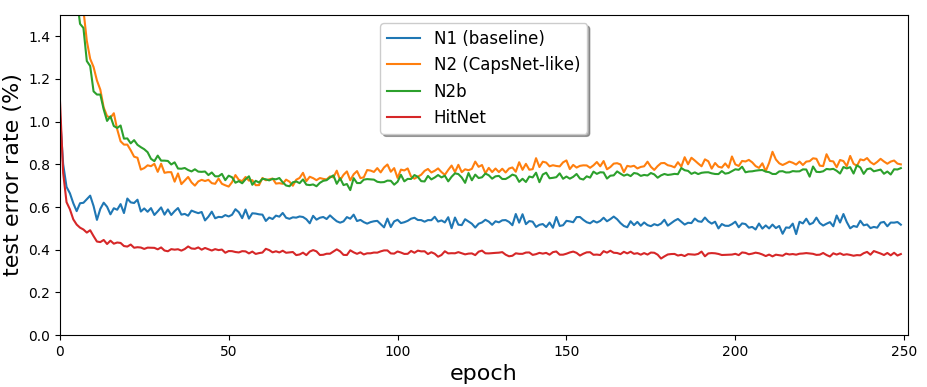}\caption{Evolution of the average test error rate on MNIST over the $20$
runs of each network as a function of the epochs, with a constant
learning rate. The superiority of $\protect\ourModel$ can be seen.
The convergence is also finer in that case, in light of the weaker
oscillations from one epoch to the next (lower irregularity).\label{fig:firstcompar-1}}
\end{figure}

The results throughout the epochs are plotted in Figure~\ref{fig:firstcompar-1}.
The error rates of the four models are reported in Table~\ref{tab:performances}
and, as it can also be seen in Figure~\ref{fig:firstcompar-1}, they
clearly indicate that the centripetal approach of $\ourModel$ is
better suited than a centrifugal approach, regardless of the activation
used. This observation is confirmed if the squashing function is used
in $\ourModel$, in which case a test error rate of about $0.40\%$
is obtained.

In Table~\ref{tab:performances}, the column ``Standard deviation
(Std)'' represents the variability that is obtained in the error
rate among the $20$ runs of a given network. The column ``Irregularity''
relates to the average (over the $20$ runs) of the standard deviation
of the last $100$ error rates recorded for each run. A low irregularity
represents a more ``natural'' convergence of the runs of a given
network since it measures the variability of the error rate within
a run. This indicator makes sense in the context of a constant learning
rate and no overfitting, as observed with $\ourModel$. 

These two metrics both indicate superior performances when using $\ourModel$,
in the sense that there is an intrinsically better natural convergence
associated with the centripetal approach (lower irregularity) and
more consistent results between the runs of a same model (lower standard
deviation). Let us note that all the runs of $\ourModel$ converged
and no overfitting is observed. The question of the convergence is
not studied in \cite{Sabour2017Dynamic} and the network is stopped
before it is observed diverging in \cite{Nair2018Pushing}.

Then, we run the same experiments but with a decreasing learning rate,
to see how the results are impacted when the convergence is forced.
The learning rate is multiplied by a factor $0.95$ at the end of
each epoch. As a result, the networks stabilize more easily around
a local minimum of the loss function, improving their overall performances.
It can be noted that $\ourModel$ is less impacted, which indicates
that $\ourModel$ converges to similar states with or without decreasing
learning rate. The conclusions are the same as previously: $\ourModel$
performs better. Let us note that in this case, the irregularity is
not useful anymore. We replace it by the best error rate obtained
for a converged run of each type of network.

\begin{table}
\centering{}{\small{}}%
\begin{tabular}{|c|c|c|c||c|c|c|}
\cline{2-7} 
\multicolumn{1}{c|}{} & \multicolumn{3}{c||}{{\small{}Constant learning rate}} & \multicolumn{3}{c|}{{\small{}Decreasing learning rate}}\tabularnewline
\hline 
{\small{}Network} & {\small{}Err. rate \%} & {\small{}Std ($\times10^{-2}$)} & {\small{}Irreg. ($\times10^{-2}$)} & {\small{}Err. rate \%} & {\small{}Std ($\times10^{-2}$)} & {\small{}Best \%}\tabularnewline
\hline 
{\small{}Baseline} & {\small{}$0.52$} & {\small{}$0.060$} & {\small{}$0.060$} & {\small{}$0.42$} & {\small{}$0.027$} & {\small{}$0.38$}\tabularnewline
\hline 
{\small{}CapsNet-like} & {\small{}$0.79$} & {\small{}$0.089$} & {\small{}$0.053$} & {\small{}$0.70$} & {\small{}$0.076$} & {\small{}$0.53$}\tabularnewline
\hline 
{\small{}N2b} & {\small{}$0.76$} & {\small{}$0.072$} & {\small{}$0.048$} & {\small{}$0.72$} & {\small{}$0.074$} & {\small{}$0.62$}\tabularnewline
\hline 
{\small{}$\ourModel$} & {\small{}$\mathbf{0.38}$} & \textbf{\small{}$\mathbf{0.033}$} & \textbf{\small{}$\mathbf{0.025}$} & \textbf{\small{}$\mathbf{0.36}$} & \textbf{\small{}$\mathbf{0.025}$} & \textbf{\small{}$\mathbf{0.32}$}\tabularnewline
\hline 
\end{tabular}\caption{Performance metrics on MNIST of the four networks described in the
text, showing the superiority of $\protect\ourModel$.}
\label{tab:performances}
\end{table}

\subsection*{Comparisons with reported results of CapsNet on several datasets}

As far as MNIST is concerned, the best test error rate reported in~\cite{Sabour2017Dynamic}
is $0.25\%$, which is obtained with dynamic routing and is an average
of 3 runs only. However, to our knowledge, this result has yet to
be confirmed, as the best tentative reproductions reach error rates
which compare with our results, as shown in Table~\ref{tab:compar}.
Let us note that, in~\cite{Wan2013Regularization}, authors report
a $0.21\%$ test error rate, which is the best performance published
so far. Nevertheless, this score is reached with a voting committee
of five networks that were trained with random crops, rotation and
scaling as data augmentation processes. They achieve $0.63\%$ without
the committee and these techniques; if only random crops are allowed
(as done here with the shifts of up to 2 pixels), they achieve $0.39\%$.
It is important to underline that such implementations report excessively
long training times, mainly due to the dynamic routing part. For example,
the implementation~\cite{Guo2017CapsNetKeras} appears to be about
$13$ times slower than $\ourModel$, for comparable performances.
Therefore, $\ourModel$ produces results consistent with state-of-the
art methods on MNIST while being simple, light and fast.

The results obtained with $\ourModel$ and those obtained with CapsNet
in different sources on Fashion-MNIST, CIFAR10, SVHN, affNIST are
also compared in Table~\ref{tab:compar}. For fair comparisons, the
architecture of $\ourModel$ described in Section~\ref{subsec:Centripetal-loss-and-HitNet}
is left untouched. The results reported are obtained with a constant
learning rate and are average error rates on $20$ runs as previously.
Some comments about these experiments are given below:
\begin{enumerate}
\item Fashion-MNIST: $\ourModel$ outperforms reproductions of CapsNet except
for \cite{Guo2017CapsNetKeras}, but this result is obtained with
horizontal flipping as additional data augmentation process. 
\item CIFAR10: $\ourModel$ outperforms the reproductions of CapsNet. The
result provided in \cite{Sabour2017Dynamic} is obtained with an ensemble
of $7$ models. However, the individual performances of $\ourModel$
and of the reproductions do not suggest that ensembling them would
lead to that result, as also suggested in \cite{Xi2017Capsule}, which
reaches between $28\%$ and $32\%$ test error rates.
\item SVHN: $\ourModel$ outperforms CapsNet from \cite{Nair2018Pushing},
which is the only source using CapsNet with this dataset.
\item affNIST: $\ourModel$ outperforms the results provided in \cite{Shin2018CapsNetTensorFlow}
and even in Sabour \etal by a comfortable margin. We performed the
same experiment as the one described in \cite{Sabour2017Dynamic}.
Each image of the MNIST train set is placed randomly (once and for
all) on a black background of $40\times40$ pixels, which constitutes
the training set of the experiment. The images of the batches are
not randomly shifted of up to $2$ pixels in each direction anymore.
After training, the models are tested on affNIST test set, which consists
in affine transformations of MNIST test set. Let us note that a test
error rate of only $2.7\%$ is obtained if the images of the batches
are randomly shifted of up to $2$ pixels in each direction as for
the previous experiments.
\end{enumerate}
\begin{table}
\centering{}%
\begin{tabular}{|c|c|c|c|c|c|}
\hline 
CapsNet from & MNIST & Fashion-MNIST & CIFAR10 & SVHN & affNIST\tabularnewline
\hline 
\hline 
Sabour \etal \cite{Sabour2017Dynamic} & 0.25 & - & 10.60 & 4.30 & 21.00\tabularnewline
\hline 
Nair \etal \cite{Nair2018Pushing} & 0.50 & 10.20 & 32.47 & 8.94 & -\tabularnewline
\hline 
Guo \cite{Guo2017CapsNetKeras} & 0.34 & 6.38 & 27.21 & - & -\tabularnewline
\hline 
Liao \cite{Liao2018CapsNet} & 0.36 & 9.40 & - & - & -\tabularnewline
\hline 
Shin \cite{Shin2018CapsNetTensorFlow} & 0.75 & 10.98 & 30.18 & - & 24.11\tabularnewline
\hline 
\hline 
$\ourModel$ & 0.38/0.32 & 7.70 & 26.70 & 5.50 & 16.97\tabularnewline
\hline 
\end{tabular}\caption{Comparison between the error rates (in \%) reported on various experiments
with CapsNet and $\protect\ourModel$, in which case the average results
over $20$ runs are reported.}
\label{tab:compar}
\end{table}

\subsection{Using the decoder for visualization and hybrid data augmentation}

In this section, we present some results related to Section~\ref{subsec:Prototypes,-data-generation-1}
about the uses of the decoder to build prototypes, to perform data
generation and data augmentation.

\subsubsection*{Constructing prototypes, interpreting the features and data generation}

\begin{figure}
\centering{}\includegraphics[width=0.8\textwidth]{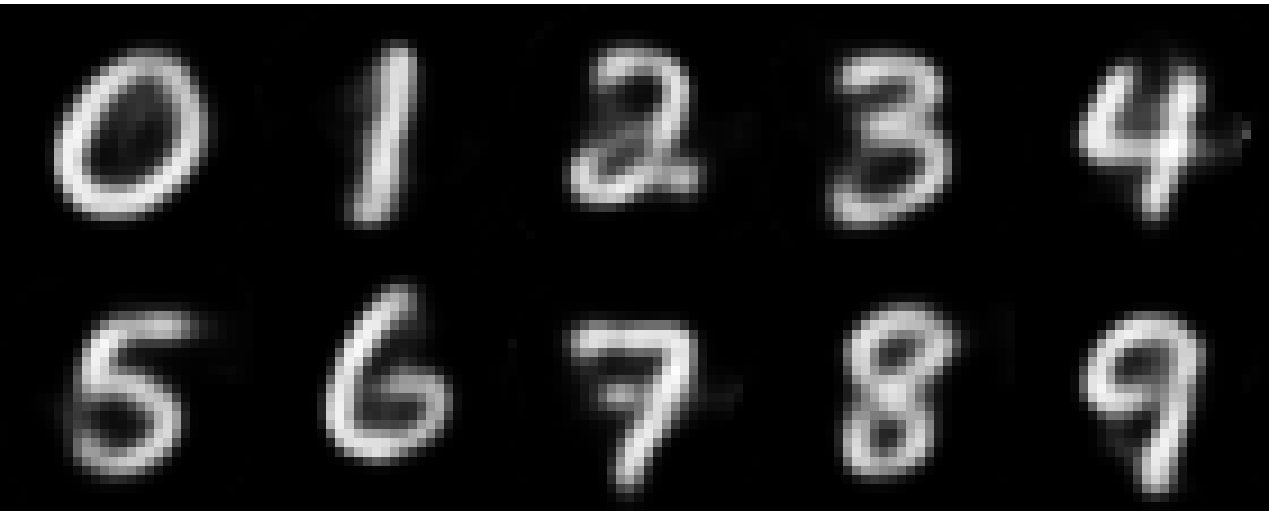}\caption{Prototypes obtained at the end of the training by feeding the decoder
with capsules of zeros except one, which is replaced by the central
capsule $\protect\centerTarget:(0.5,\,\ldots,\,0.5)$. These prototypes
can be seen as the reference images from which $\protect\ourModel$
evaluates the similarity with the input image through HoM. \label{fig:prototypes}}
\end{figure}

As mentioned in Section~\ref{subsec:Prototypes,-data-generation-1},
the centripetal approach gives a particular role to the central capsule
$\centerTarget:(0.5,...,0.5)$, in the sense that it can be used to
generate prototypes of the different classes. The prototypes obtained
from an instance of $\ourModel$ trained on MNIST are displayed in
Figure~\ref{fig:prototypes}.

It is then particularly easy to visualize what each component represents
by tweaking the components of $\centerTarget$ around $0.5$; there
is no need to distort real images as in~\cite{Sabour2017Dynamic}.
Also, in our case, as mentioned in Section~\ref{subsec:Prototypes,-data-generation-1},
the range $[0.5-2m/\sqrt{n},0.5+2m/\sqrt{n}]$ is suitable to tweak
the parameters, while these ranges are not predictable with CapsNet
and may vary from one feature to another, as it can be inferred from
\cite{Shahroudnejad2018Improved,Zhang2018CapProNet} where the capsules
of a class do not necessarily fall close to each other. The range
in question $[0.5-2m/\sqrt{n},0.5+2m/\sqrt{n}]$ is actually confirmed
by looking at the distributions of the individual features embodied
in the capsules. On MNIST, $\ourModel$ captures some features that
are positional characteristics, others can be related to the width
of the font or to some local peculiarities of the digits, as in~\cite{Sabour2017Dynamic}.
Sampling random capsules close to $\centerTarget$ for a given class
thus generates new images whose characteristics are combinations of
the characteristics of the training images. It thus makes sense to
encompass all the capsules of training images of that class in a convex
space, as done with $\ourModel$, to ensure the consistency of the
images produced, while CapsNet does not guarantee this behavior (\cite{Shahroudnejad2018Improved,Zhang2018CapProNet}).

Let us note that, as underlined in~\cite{Nair2018Pushing}, the reconstructions
obtained for Fashion-MNIST lacks details and those of CIFAR10 and
SVHN are somehow blurred backgrounds; this is also the case for the
prototypes. We believe that at least three factors could provide an
explanation: the decoder is too shallow, the size of the capsules
is too short, and the fact that the decoder has to reconstruct the
whole image, including the background, which is counterproductive.

\subsubsection*{Hybrid data augmentation}

\begin{figure}
\subfloat[]{\includegraphics[width=0.47\textwidth]{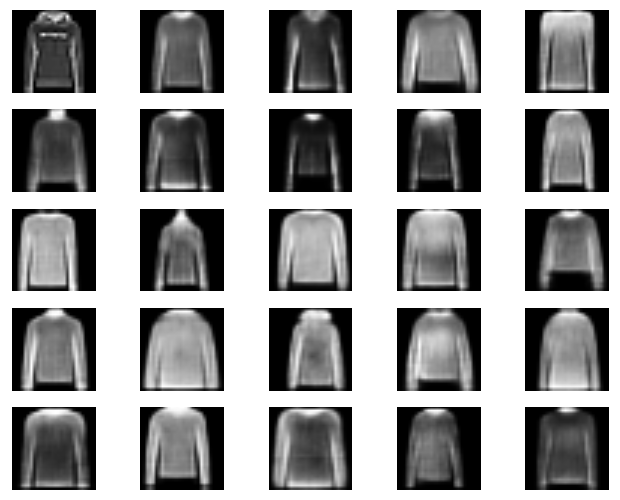}}\hfill{}\subfloat[]{\includegraphics[width=0.47\textwidth]{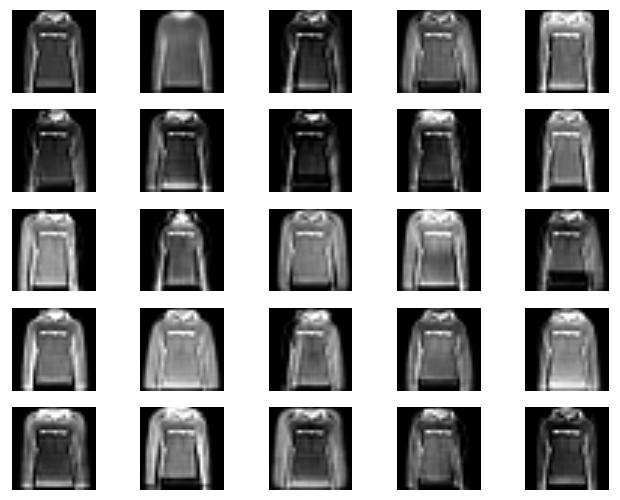}}

\subfloat[]{\includegraphics[width=0.47\textwidth]{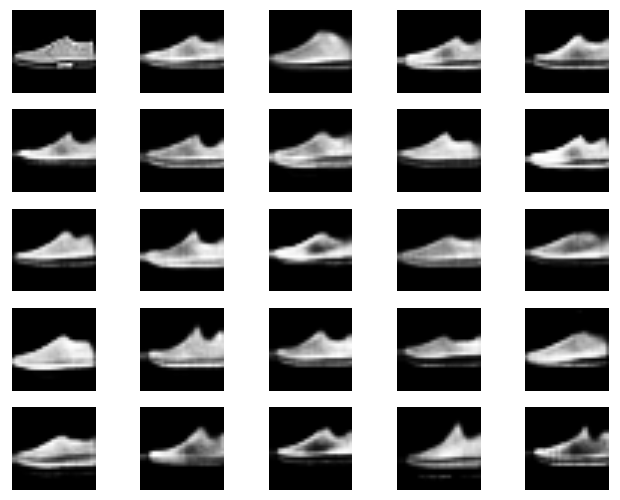}}\hfill{}\subfloat[]{\includegraphics[width=0.47\textwidth]{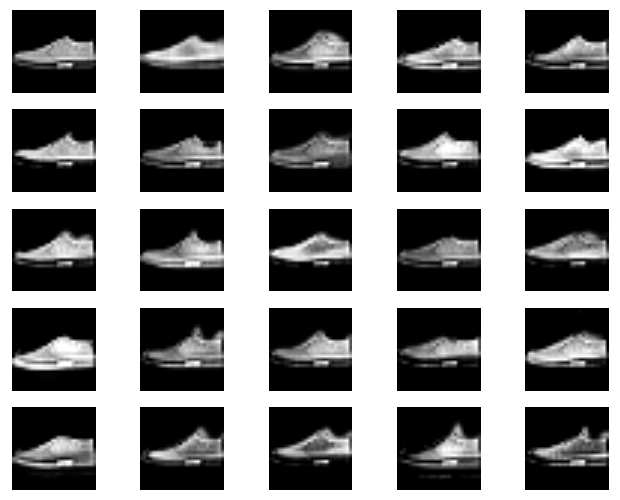}}\caption{Examples of hybrid data augmentation using the Fashion-MNIST dataset
(the resolution is increased for more visual comfort). In (a) and
(b), the top left image is an original image $\protect\trainingImage$
from the dataset and the image on its right is its corresponding reconstruction
$\protect\reconstructed{\protect\trainingImage}$ from HoM (these
images are the same in (a) and (b)). In (a), the $23$ remaining images
are modified versions of $\protect\reconstructed{\protect\trainingImage}$
obtained by tweaking the components of its capsule in HoM. These images
correspond to $23$ examples of $\protect\modified{\protect\trainingImage}$.
We can see that they do not contain any detail but rather serve as
a skeleton on which the details $\protect\trainingImage-\protect\reconstructed{\protect\trainingImage}$
should be added. Adding these details to each of them gives the $23$
remaining images of (b), thus displaying $\protect\modified{\protect\trainingImage}+\protect\trainingImage-\protect\reconstructed{\protect\trainingImage}$,
clipped to the $[0,1]$ range, which are the images generated by the
hybrid data augmentation explained in the text. The same process is
shown with a different original image $\protect\trainingImage$ in
(c) and (d). We can see that adding the details allows to generate
convincing detailed images. \label{fig:dataaug}}
\end{figure}

The quality of the data generated as described above depends on multiple
factors, such as the number of features extracted and the quality
of the decoder. Given a restricted amount of information, that is,
capsules of size $16$, the decoder can only reconstruct approximations
of initial images, which may not look realistic enough in some contexts.
In order to incorporate the details lost in the computation of HoM,
the hybrid feature-based and data-based data augmentation technique
described in Section~\ref{subsec:Prototypes,-data-generation-1}
can be applied. The importance of adding the details and thus the
benefits over the sole data generation process can be visualized in
Figure~\ref{fig:dataaug} with the FashionMNIST dataset. 

The classification performances can be marginally increased with this
data augmentation process as it appeared that networks trained from
scratch on such data (continuously generated on-the-fly) perform slightly
better than when trained with the original data. On MNIST, the average
error rate on $20$ models decreased to $0.33\%$ with a constant
learning rate and to $0.30\%$ with a decreasing learning rate. In
our experiments, $3$ of these models converged to $0.26\%$, one
converged to $0.24\%$. Some runs reached $0.20\%$ test error rate
at some epochs. With a bit of luck, a blind selection of one trained
network could thus lead to a new state of the art, even though it
is known that MNIST digits classification results will probably not
reach better performances due to inconsistencies in the test set.
The average test error rate on Fashion-MNIST decreases by $0.2\%$,
on CIFAR10 by $1.5\%$ and on SVHN by $0.2\%$. These results could
presumably be improved with more elaborate feature maps extractors
and decoders, given the increased complexity of these datasets. The
data augmentation is not performed on affNIST since the purpose of
this dataset is to measure the robustness of the network to (affine)
transformations.

\begin{table}
\centering{}%
\begin{tabular}{|c|c|c|c|c|}
\hline 
 & MNIST & Fashion-MNIST & CIFAR10 & SVHN\tabularnewline
\hline 
\hline 
$\ourModel$ & 0.38 & 7.70 & 26.70 & 5.50\tabularnewline
\hline 
$\ourModel$ with DA & 0.33 & 7.50 & 25.20 & 5.30\tabularnewline
\hline 
$\ourModel$ with DA and dlr (best) & 0.24 & - & - & -\tabularnewline
\hline 
\end{tabular}\caption{Test error rates obtained with the hybrid data augmentation (DA) process,
showing slightly better performances. On MNIST, we obtained state-of-the-art
performance with a decreasing learning rate.}
\end{table}

\subsection{Analyzing a training set and detecting possibly mislabeled images
with ghost capsules}

In this section, we illustrate the use of ghost capsules described
in Section~\ref{subsec:Ghost-capsules} to analyze the training set
of MNIST and detect potentially erroneous labels. We train $\ourModel$
$20$ times with GC, which gives us $20$ models, and we examine the
$20$ GC\footnote{For lighter notations, we note GC both the singular and the plural
form of ghost capsule(s), which will not be misleading in the following
given the context.} associated with each image by these models. 

\subsubsection*{Agreement between the models}

We first study the agreement between the models about the $20$ GC
classes selected for each image. For that purpose, we examine the
distribution of the number of different classes $\numberOfFSClasses$
given to the $20$ GC of each image. It appears that the $20$ models
all agree ($\numberOfFSClasses=1$) on the same GC class for $28\%$
of the training images, which represents $16.859$ images. 

To refine our analysis, we now focus on the $16.859$ images that
have all their GC in the same class ($\numberOfFSClasses=1)$ and
that potentially make some hits. These images have a pair $\text{(true class, GC class)}$
and their distribution suggests that some pairs are more likely to
occur, such as $(3,5)$ for $2333$ images, $(4,9)$ for $2580$ images,
$(7,2)$ for $1360$ images, $(9,4)$ for $2380$ images, which gives
a glimpse of the classes that might be likely to be mixed up by the
models. Confusions may occur because of errors in the true labels,
but since these numbers are obtained on the basis of an agreement
of $20$ models trained from the same network structure, this may
also indicate the limitations of the network structure itself in its
ability to identify the different classes. However, a deeper analysis
is needed to determine if these numbers indicate a real confusion
or not, that is, which of them make hits and which do not.

So far, we know that for these $16.859$ images, the $20$ models
agree on one GC class ($\numberOfFSClasses=1)$. We can examine if
the $20$ models agree on their distance from $\centerTarget$ for
their class. For that purpose, for each image, we compute the mean
and the standard deviation of its $20$ GC distances from $\centerTarget$.
An interesting observation is that when the mean distance gets closer
to the hit zone threshold (which is $m=0.1$), then the standard deviation
decreases, which indicates that all the models tend to agree on the
fact that a hit is needed. This is particularly interesting in the
case of mislabeled images (examples are provided hereafter, in Figure~\ref{fig:all-monsters}).
Indeed, if an image of a ``4'' is labeled as ``5'' (confusion),
not only are the $20$ models likely to provide $20$ GC with the
same class (``4'' here), but they will also all make a hit in that
class ``4'' and thus identify it as a highly doubtful image.

\subsubsection*{Identification of doubtful images}

We can now narrow down the analysis to the images that are the most
likely to be mislabeled, that is, those with a mean distance smaller
than $m$; there are only $71$ such images left. The associated pairs
$\text{(true class, GC class)}$ are given in Table~\ref{tab:dist-fs-true-hits}.
From the expected confusions mentioned above $(3,5)$, $(4,9)$, $(7,2)$,
$(9,4)$, we can see that $(7,2)$ is actually not so much represented
in the present case, while the pairs $(1,7)$ and $(7,1)$ subsisted
in a larger proportion, and that the pairs $(4,9)$ and $(9,4)$ account
for almost half of the images of interest.

\begin{table}
\centering{}%
\begin{tabular}{cc|cccccccccc|c}
 &  & \multicolumn{10}{c|}{True class} & \multirow{2}{*}{Total}\tabularnewline
 &  & 0 & 1 & 2 & 3 & 4 & 5 & 6 & 7 & 8 & 9 & \tabularnewline
\hline 
\multirow{10}{*}{GC class} & 0 & 0 & 0 & 0 & 0 & 0 & 0 & 0 & 0 & 0 & 0 & 0\tabularnewline
 & 1 & 0 & 0 & 0 & 0 & \textcolor{blue}{2} & 0 & 0 & \textcolor{blue}{6} & 0 & 0 & 8\tabularnewline
 & 2 & 0 & 0 & 0 & 0 & 0 & 0 & 0 & \textcolor{blue}{1} & 0 & 0 & 1\tabularnewline
 & 3 & 0 & 0 & 0 & 0 & 0 & \textcolor{blue}{2} & 0 & 0 & 0 & 0 & 2\tabularnewline
 & 4 & 0 & 0 & 0 & 0 & 0 & 0 & \textcolor{blue}{1} & 0 & 0 & \textcolor{blue}{20} & 21\tabularnewline
 & 5 & 0 & 0 & 0 & \textcolor{blue}{8} & 0 & 0 & \textcolor{blue}{1} & 0 & 0 & 0 & 9\tabularnewline
 & 6 & 0 & 0 & 0 & 0 & 0 & \textcolor{blue}{2} & 0 & 0 & 0 & 0 & 2\tabularnewline
 & 7 & 0 & \textcolor{blue}{7} & \textcolor{blue}{2} & 0 & \textcolor{blue}{1} & 0 & 0 & 0 & 0 & \textcolor{blue}{2} & 12\tabularnewline
 & 8 & 0 & 0 & 0 & \textcolor{blue}{1} & 0 & 0 & 0 & 0 & 0 & 0 & 1\tabularnewline
 & 9 & 0 & 0 & 0 & \textcolor{blue}{1} & \textcolor{blue}{13} & 0 & 0 & \textcolor{blue}{1} & 0 & 0 & 15\tabularnewline
\hline 
\multicolumn{2}{c|}{Total} & 0 & 7 & 2 & 10 & 16 & 4 & 2 & 8 & 0 & 22 & 71\tabularnewline
\end{tabular}\caption{Distribution of pairs $\text{(true class, GC class)}$ for MNIST training
images having a unique GC class and their mean distance from $\protect\centerTarget$
smaller than the hit zone threshold.}
\label{tab:dist-fs-true-hits}
\end{table}

\begin{figure}
\begin{centering}
\subfloat[]{\includegraphics[height=0.6\textheight]{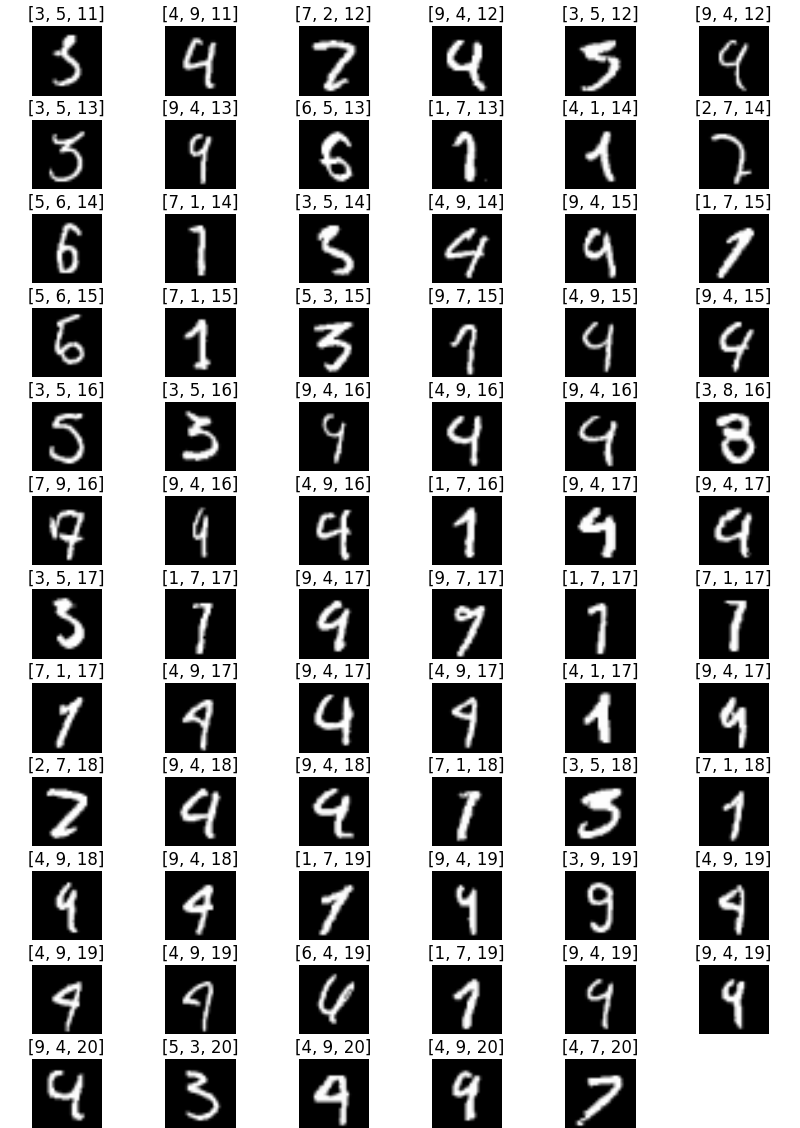}}
\par\end{centering}
\centering{}\subfloat[]{\includegraphics[height=0.2\textheight]{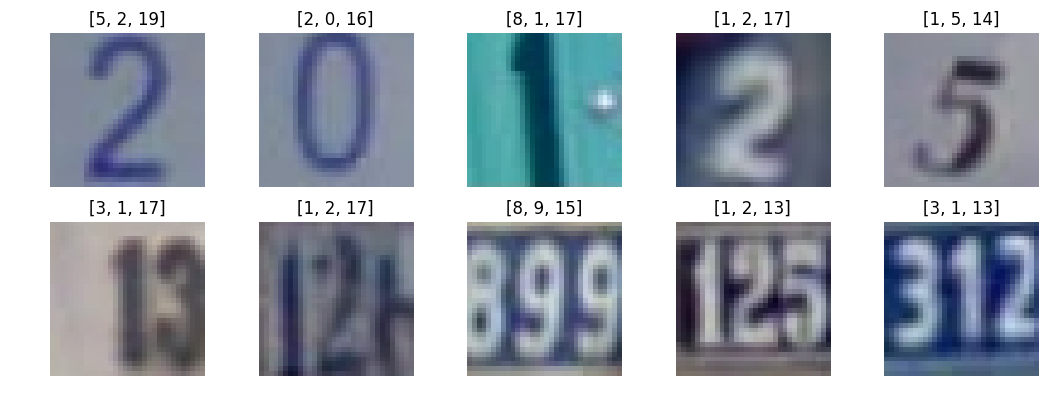}}\caption{(a) The $71$ images (resolution is increased for more visual comfort)
of MNIST training set whose $20$ GC have a mean distance from $\protect\centerTarget$
smaller than the hit zone threshold, sorted by an increasing number
of hits. The three numbers in brackets indicate respectively the true
label, the GC class, the number of hits in the GC class. (b) A selection
of images of SVHN obtained following the same process with ghost capsules.
In the first row, the images are clearly mislabeled. In the second
row, the images deserve several labels.\label{fig:all-monsters}}
\end{figure}

The last refinement that we make is a look at the number of hits among
the $20$ GC of these $71$ images. We know that their mean distance
from $\centerTarget$ is smaller than $m$, but neither this mean
distance nor the standard deviation clearly indicate how many of the
$20$ models actually make a hit for these $71$ images. It appears
that all these images have at least $55\%$ $(11/20)$ of their GC
in the hit zone and that more than $75\%$ $(55/71)$ of the images
have a hit for at least $75\%$ $(15/20)$ of the models, which indicates
that when the mean distance is smaller than $m$, it is the result
of a strong agreement between the models. Finally, the $71$ images,
sorted by number of hits, are represented in Figure~\ref{fig:all-monsters}.
The true label, the GC class, and the number of hits are indicated
for each image. Some of these images are clearly mislabeled and some
are terribly confusing by looking almost the same but having different
labels, which explains the GC class selected and the number of hits
obtained. While pursuing a different purpose, the DropMax technique
used in~\cite{Lee2017DropMax} allowed the authors to identify ``hard
cases'' of the training set which are among the $71$ images represented
in Figure~\ref{fig:all-monsters}. 

Following the same process as above, we use ghost capsules with SVHN.
Again, we are able to detect images that are clearly mislabeled, as
seen in Figure~\ref{fig:all-monsters}. In addition, in this case,
the use of ghost capsules allows us to detect images that deserve
multiple labels since multiple digits are present in the images, as
seen in Figure~\ref{fig:all-monsters}.

Ghost capsules could be used in other contexts. For example, we could
imagine that a human annotator hesitates between two classes in the
labelling process because the image is confusing. The annotator chooses
one class and the ghost capsule deals with the confusion by potentially
choosing the other class without penalizing the network. In another
scenario, several annotators may give two different labels for the
same image. We could handle this potential conflict with ghost capsules,
by first attributing the image one of the labels suggested by the
annotators (\eg the most suggested) then training the network and
finally checking whether the ghost capsule of that image is associated
with the other label suggested and, if so, whether it makes a hit
or not.

\section{Conclusion}

We introduce $\ourModel$, a deep learning network characterized by
the use of a Hit-or-Miss layer composed of capsules, which are compared
to central capsules through a new centripetal loss. The idea is that
the capsule corresponding to the true class has to make a hit in its
target space, and the other capsules have to make misses. The novelties
reside in the reinterpretation and in the use of the HoM layer, which
provides new insights on how to use capsules in neural networks. Besides,
we present two additional possibilities of using $\ourModel$. In
the first one, we explain how to build prototypes, which are class
representatives, how to deform them to perform data generation, and
how to set up a hybrid data augmentation process. This is done by
combining information from the data space and from the feature space.
In the second one, we design ghost capsules that are used to allow
the network to alleviate the loss of capsules related to plausible
alternative classes. 

In our experiments, we demonstrate that $\ourModel$ is capable of
reaching state-of-the-art performances on MNIST digits classification
task with a shallow architecture and that $\ourModel$ outperforms
the results reproduced with CapsNet on several datasets, while being
at least $10$ times faster. The convergence of $\ourModel$ does
not need to be forced by a decreasing learning rate mechanism to reach
good performances. $\ourModel$ does not seem to suffer from overfitting,
and provides a small variability in the results obtained from several
runs. We also show how prototypes can be built as class representatives
and we illustrate the hybrid data augmentation process to generate
new realistic data. This process can also be used to marginally increase
classification performances. Finally, we examplify how the ghost capsules
help identifying suspicious labels in the training set, which allows
to pinpoint images that should be considered carefully in the training
process.

\subsubsection*{Future work}

As far as the classification performances are concerned, one of the
main advantages of the HoM layer is that is can be incorporated in
any other network. This implies that the sub-part of $\ourModel$
used to compute the feature maps that are fully connected to HoM can
be replaced by more elaborate networks to increase the performances
on more complex tasks.

In a similar way, the prototypes and all the reconstructions made
by the decoder could be improved by using a more advanced decoder
sub-network and capsules with more components. In real-life cases
such as CIFAR10 and SVHN, it could also be useful to make a distinction
between the object of interest and the background. For example, features
designed to reconstuct the background only could be used. If segmentation
masks are available, one could also use the capsules to reconstruct
the object of interest in the segmented image, or simply the segmentation
mask. One could also imagine to attach different weights to the features
captured by the capsules, so that those not useful for the classification
are used in the reconstruction only. The flexibility of HoM allows
to implement such ideas easily.

Regarding ghost capsules, they could be used to perform a built-in
top-k classification, by using $k$ ghost capsules in the training
process. In a binary classification task, they could be used only
after a given number of epochs, which would make more sense than using
them at the beginning of the training. The ghost capsule could also
generate some loss but with a given probability to take into account
the fact that it does not correspond to the true class and that is
should be penalized in some way.

Comparing the results on other benchmark datasets would help promoting
$\ourModel$ in a near future as well.

\subsubsection*{Acknowledgements}

We are grateful to M. Braham for introducing us to the work of Sabour et al. and for all the fruitful discussions. 

This research is supported by the DeepSport project of the Walloon
region, Belgium. A. Cioppa has a grant funded by the FRIA, Belgium.
We also thank NVIDIA for the support.

\end{document}